Author's accepted version.
Ngo, V. Z. H., Patel, R., Ramchurn, R., Chamberlain, A., & Kucukyilmaz, A., "Dancing with a Robot: An Experimental Study of Child-Robot Interaction in a Performative Art Setting," in Proceedings of the 16th International Conference on Social Robotics (ICSR), Odense, Denmark, 23-26 October 2024.

# Dancing with a Robot: An Experimental Study of Child-Robot Interaction in a Performative Art Setting

Victor Zhi Heung Ngo[1][0009-0003-0805-5292], Roma Patel[0000-0003-2075-2560], Rachel Ramchurn, Alan Chamberlain[0000-0002-2122-8077], Ayse Kucukyilmaz[0000-0003-3202-6750]

[1] School of Computer Science, University of Nottingham, Nottingham, United Kingdom
{victor.ngo, alan.chamberlain, ayse.kucukyilmaz}@nottingham.ac.uk
{roma, rachel}@makersofimaginaryworlds.co.uk

**Abstract.** This paper presents an evaluation of 18 children's in-the-wild experiences with the autonomous robot arm performer NED (Never-Ending Dancer) within the Thingamabobas installation, showcased across the UK. We detail NED's design, including costume, behaviour, and human interactions, all integral to the installation. Our observational analysis revealed three key challenges in child-robot interactions: 1) Initiating and maintaining engagement, 2) Lack of robot expressivity and reciprocity, and 3) Unmet expectations. Our findings show that children are naturally curious, and adept at interacting with a robotic art performer. However, our observations emphasise the critical need to optimise human-robot interaction (HRI) systems through careful consideration of audience's capabilities, perceptions, and expectations, within the performative arts context, to enable engaging and meaningful experiences, especially for young audiences.

## 1 Introduction

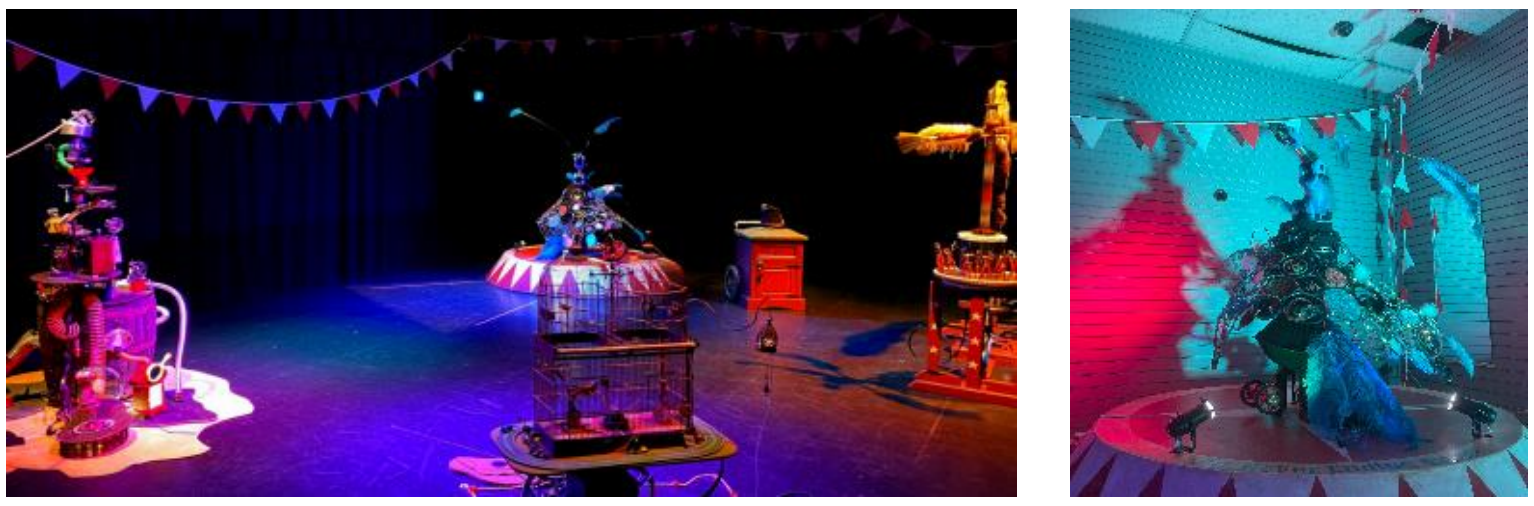

**Fig. 1.** (left) NED is part of the Thingamabobas installation [1], which offers a playful sensory experience, where participants engage with a circus troupe of performative mechanical and kinetic structures. (right) The Never-Ending Dancer (NED) is presented on a theatrical stage at the National Festival of Making, Blackburn.

With the rapid advancements in AI over recent years, robotics is evolving beyond conventional, accepted practical applications and making its mark in the realm of the performative arts. The “Thingamabobas" installation [1] is an artist-led installation project by Makers of Imaginary Worlds (MOIW)[1]. It embodies the fusion of technology and art in a child-robot interaction setting (Fig. 1 (left)), through the creation of a dancing autonomous robot arm performer called the NED, Never-Ending Dancer, shown in Fig. 1 (right). As with any artistic performance, robotic performative art relies heavily on audience interaction, making user engagement a critical element.

This paper presents data and insights from two in-the-wild experiments [2], in which 18 children interacted with NED over a period of three weeks, totalling more than four hours of engagement. Through our observational analysis of the participant’s interactions with NED and a thematic analysis of the semi-structured interviews conducted shortly after their interactions, we explore the following research questions:

RQ1: How do children engage with a performance art robot, and what causes disengagement in this?

RQ2: What are the children’s perceptions of their specific interaction with NED?

Based on our findings, we reflect strategies to enhance engagement in child-robot interaction (CRI).

This paper is organised as follows: Section 2 discusses the current state of engagement in CRI in performative art settings, Section 3 details the design of NED, Section 4 details the study procedure, Section 5 explains the methodology, Section 6 presents a discussion on the interaction videos and interview findings, Section 7 concludes this paper’s findings, Section 8 reviews the study’s limitations, and section 9 closes this paper with reflections on future work.

## 2 Related Work

Children’s perceptions towards robots in performative art settings have been observed to vary considerably, with factors such as the robot's appearance, behaviour and the context of the interaction identified as influencing these perceptions [3, 4]. It has been demonstrated that children can form social bonds (relationships) with robots, perceiving them as social agents rather than mere machines [5]. Studies have shown that the integration of robots in artistic performances can strengthen the bond between performers and the audience, leading to higher engagement levels [6].

Understanding what factors influence these bonds has the potential to enhance engagement, and in turn enable better methods for child-robot interaction. This is a crucial step for developing effective CRI methods in a world where children are increasingly anticipated to engage with social robots [7].

Research in CRI often focuses on educational [8, 9] or therapeutic contexts [10], with limited exploration of how children perceive and interact with robots in dynamic in-the-wild artistic performances. In an artistic context, Silvera-Tawil et al. [11] investigated quantitative differences between adults and children in the way they interact with an interactive mobile robot art structure in a museum. They studied the speed of movement to and from the robot, distance to the robot, time spent in various proxemics distances to the robot, and relative gaze. They highlight that children tend to interact within a closer proximity to the robot compared to adults, however, the children's perception of the interaction, and the quality of the engagement was not explored.

[1] https://makersofimaginaryworlds.co.uk/

Performance art robots, designed to entertain and educate, provide a distinctive context for studying engagement. Leite et al. [12] found that narrative elements, emotional expressions, and interactive storytelling significantly enhance engagement. The immersive nature of performance art, combined with a robot's ability to respond to human cues, can create dynamic interactions that captivate especially the younger audiences.

Additionally, what motivates children to engage in interactions is equally important as to how they engage. Curiosity is understood to be a fundamental motivator behind children's engagement with autonomous systems. Gordon et al. [13] highlighted that robots designed with features that elicit curiosity, such as unpredictable behaviours or problem-solving challenges, tend to sustain children's attention for longer periods of time, emphasising the significance of novelty and interactivity in sustaining children's interest. This may suggest that phenomena such as novelty effect, should be something that is utilised and designed for to enhance user engagement in CRI [14]. Further research by Westlund et al. [15] indicated that children's curiosity is at its highest when robots present scenarios that require exploration and discovery. As emphasised in [15], the role of emotional intelligence in robots, where the ability to recognise and respond to children's emotions, significantly boosts engagement.

However, disengagement can also occur due to several factors, including technical issues, lack of novelty, and perceived predictability. Kennedy et al. [16] identified that technical malfunctions or lag in robot responses lead to frustration and reduced engagement. Additionally, children quickly lose interest if the robot's behaviour becomes repetitive or fails to offer new challenges [17]. Furthermore, a mismatch between the robot's capabilities and the child's expectations can result in disinterest and ultimately disengagement.

Given the importance of engagement in CRI, and the lack of qualitative approaches for in-the-wild experiments in this area, we report findings from two installations featuring NED interacting with 18 children. By observing and analysing the interactions between children and robots in naturalistic settings, we aim to gain a deeper understanding of how to design more engaging, hence effective, child-robot interactions in performative art contexts.

## 3 The Design of NED, the Never-Ending Dancer

The Thingamabobas [1] is an interactive performance created in Spring 2021 by a team of artists, technologists, a choreographer, and roboticists, and has been touring the UK ever since. This evolving project integrates new features based on experiences from each installation. This paper presents observations from installations at two venues.

Aimed at young audiences, Thingamabobas features NED, a dancing autonomous robotic arm adorned with feathers and an intricate metal skirt (See Fig. 1). Fig. 2 shows the concept design of NED as developed by the artists. Using facial recognition, NED detects and interacts with children through dance. This multi-component artistic piece incorporates costume design, musical accompaniment, and storytelling, redefining the perception of the robotic arm. NED's design and components, detailed in later sections, illustrate how fiction can help audiences suspend disbelief and immerse themselves in a new reality.

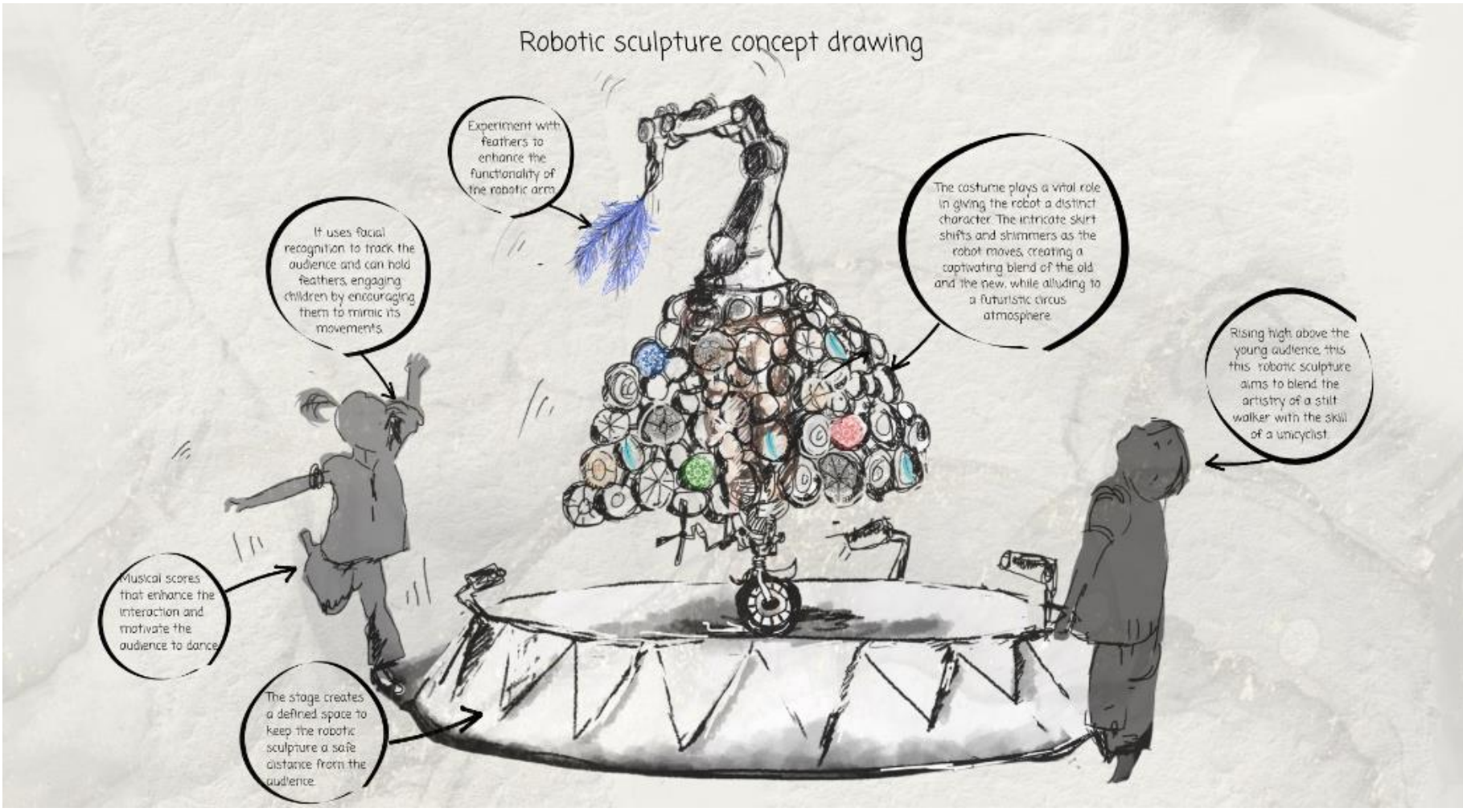

**Fig. 2.** A concept drawing of NED produced by Makers of Imaginary Worlds, 2021. https://makersofimaginaryworlds.co.uk/tas-artist-residency/

**Robotic Platform.** The Niryo NED robotic arm was used as the body of NED. This 6-axis collaborative robot arm platform is designed for education and research and is compatible with Ubuntu (version 18.04) and ROS (Melodic). The choice of an arm was intentional, as these robots typically have a functional design devoid of humanoid features or affective capabilities. We aimed to transform the robot arm into a playful kinetic sculpture that defies expectations, offering a creative reinterpretation. The robot platform was selected for its ability to perform choreographed movements with six degrees of freedom, as well as its appropriate size and cost for the research project.

**Costume Design.** NED's costume was designed to resemble a tropical, mechanical, hybrid bird-like creature, with long blue feathers attached to the arm to extend its reach, and an intricate metal skirt that swayed as the robot followed an audience member. The robot's exotic design concept is complemented by the bright blue colour in the base and rotating joint housings, which purposefully contrast with the other mechanical creations in the circus troupe of the Thingamabobas.

**Soundtrack.** NED's performance was accompanied by three musical scores intended to encourage the audience to dance. Each score was composed purposefully for the Thingamabobas installation.

**Sensors.** NED was initially equipped with a single Intel RealSense D415 depth camera, positioned front facing under the robot base. To improve tracking capabilities when the audience moves behind the robot, the system was later upgraded to use three Intel RealSense D435 depth cameras, which were positioned at the front, left rear, and right rear of the robot, all located under the robot base. This upgrade was also necessary due to a lighting-related issue, and in tandem with requests from the artists regarding the design and interaction possibilities of the robot.

**Interaction Environment.** The initial system's interaction area, deployed at the National Festival of Making, was defined by the Intel RealSense D415 depth camera's 65° FOV and a maximum distance of 4m. The upgraded system, deployed at Mansfield Museum, expanded this area using three Intel RealSense D435 depth cameras, achieving a combined 261° FOV with the same maximum distance of 4m. NED was centrally positioned in the second venue to accommodate the increased interaction range.

**Audience Detection Methods.** Facial detection on NED utilised the ROS package face_detector [18] to identify a single audience member to dance with, tested under lab conditions and used in past tours of the Thingamabobas installation. At Mansfield Museum, the upgraded system replaced audience detection with pose estimation using the YOLOv8 model [19]. Three Intel RealSense D435 cameras expanded the interaction area to a combined FOV of 261°. To improve pose estimation in low light, depth and BGR images were merged into a single frame, similar to QR code binarization [20]. Depth images in 8-bit MONO format were filtered for a 4m range and overlaid onto an 8-bit BGR image to highlight the person or object within the interaction area.

**Robot Interaction Design.** The robot behaviour was designed to estimate the distance of audience members using the Intel RealSense D415 depth camera, and depending on their distance, the robot would enter one of three distinct states; Sleep, Search, and Interact. The respective distances for the robot states were Sleep (4 metres or greater), Search (3-3.9 metres), and Interact (0.1-2.9 metres). The original intent for the three states, as envisioned by the artists, are described as:

- **Sleep** – The robot arm is folded in a downwards position, similar to a resting swan, implying that the robot is sleeping. Audience members recognise that the robot needs waking up by them.
- **Search** – The robot arm appears to look outwards, scanning the surrounding area, designed to entice the audience to come closer.
- **Interact** – The robot arm follows the audience member as they move around the stage, provoking engagement.

**Performance and narrative.** Visitors of the Thingamabobas installation were introduced to the experience by the MOIW artists or an informed member of venue staff, in a purposefully theatrical manner. A prerecorded actor was shown on a display at the entrance to the installation, providing a story and context to the creations, effectively acting as a “Thingamabobas Wrangler”. The narrative was designed to encourage children and adults to be subject to a willing suspension of disbelief and allow themselves to become immersed in the fictional reality that is Thingamabobas. Visitors were whimsically guided from the entrance of the installation, through to each of the four mechanical circus creations, with both children and adults alike reacting positively.

# 4 Interaction Experiments

## 4.1 Study Venues

The study was conducted over a three-week period between July and September 2023, across two venues: the National Festival of Making in Blackburn and Mansfield Museum in Mansfield, UK. National Festival of Making is a prominent two-day event that attracts visitors, creators, and designers from across the country. Mansfield Museum, located in Nottinghamshire and managed by the Mansfield District Council, is a local authority museum that showcases natural history and hosts artistic installations. The study spanned two days at the National Festival of Making and fifteen days at Mansfield Museum.

## 4.2 Participants

Throughout the study, 18 participants were recruited, 13 from the National Festival of Making, and 5 from Mansfield Museum. Eight participants identified as female, while 10 participants identified as male. The youngest participant was 5 years old, and the oldest was 11 years old. At the National Festival of Making, participant ages ranged from 7 to 11 years old ($M = 8.4$, $SD = 1.1$, $N = 13$). At Mansfield Museum, participant ages ranged from 5 to 9 years old ($M = 7.6$, $SD = 1.5$, $N = 5$).

The extensive publicity of the National Festival of Making facilitated the recruitment of a larger number of participants compared to Mansfield Museum. The festival was held on a weekend in July, during the peak of summer holiday for children in the UK. In contrast, the Thingamabobas tour at Mansfield Museum was scheduled for the last two weeks of summer holiday, which included a bank holiday weekend, and the final week coinciding with the return to school. Despite its extended duration, the timing likely contributed to the lower than anticipated number of participants.

## 4.3 Ethics Approval Statement

The studies involving human participants were reviewed and approved by the University of Nottingham’s School of Computer Science Research Ethics Committee (reference no. CS-2021-R34) and adhered to the University of Nottingham Code of Research Conduct and Research Ethics [21]. The procedures included providing a plain language statement and written parental consent form to all participants, informing them of the aims and anticipated outcomes of the research. Participants' right to abstain or withdraw from the project at any time was upheld. Both raw and analysed data were anonymised and stored in a secure, project-specific data system.

## 4.4 Sessions

At the entrance to the installation area, visitors were invited to take part in the study by the lead researcher, and informed consent was provided by both the participant, and their parent/guardian/carer. Once consent to participate was given, participants were introduced to NED and given instructions to interact freely with NED in a well-defined space without physically touching the robot. MOIW artists, the lead researcher and venue staff were present to prevent participants or children from attempting to touch the robot, in respect to visitor health and safety. In addition, a rope barrier, at the discretion of the artists, was implemented at the second venue to physically prevent visitors from touching the robot. Participant interactions with the robot lasted for 4.4 minutes on average, with participant interaction durations ranging from 1.5 to 8.75 minutes (M = 4.4, SD = 0.001, N = 18). Participant interviews lasted for 7.9 minutes on average, with an interview duration range from 2.9 to 17.8 minutes (M = 7.9, SD = 0.002, N = 18).

### 4.5 Data Collection Methods

Participants' interaction data were collected using video (action camera, body camera), audio (dictaphones), and field notes by the lead researcher. After each session, participants were interviewed about their interactions with the robot, capturing their thoughts and feelings. In total, 18 semi-structured interviews were conducted. Of the 18 participants, 2 participants completed approximately half of the interview before being unable to continue. Interviews were recorded using dictaphones, and a body camera to capture any useful gestures or movements for further analysis.

## 5 Methodology

### 5.1 Data Analysis Procedure

Data preparation and analysis was conducted initially by the lead researcher and validated by the wider research team. The participant interaction videos were manually annotated, using ELAN 6.7 [22], noting interaction modalities, distances to NED and other actors, displayed emotions, verbal comments, engagement levels, and robot state. These annotations were exported as PDF to NVivo 14 for coding.

Semi-structured interviews were subject to thematic analysis [23], manually transcribed, and then coded using NVivo 14. We follow an inductive coding approach, allowing codes to be developed from the data and then clustered into themes, appropriate for the exploratory nature of the research.

### 5.2 Participant Interviews

Participants were interviewed shortly after their session with NED to capture immediate reflections. Interviews took place adjacent to the interaction space with NED in view, aiding children in providing accurate and detailed responses. All interviews were conducted in English, with participants being fluent or proficient in the language.

The interview structure was formed around five topics: 1) Preconceptions about robots, 2) Perceptions of NED and its potential purpose, 3) NED's capacity to think and feel, 4) Merits of NED, and 5) Future design considerations. For each topic, a starting question was prepared, with additional questions posed by the researcher, based on the participant's responses.

# 6 Findings and Discussion

In this section we discuss the observations from the participants' interaction videos, with the aim of forming an understanding of the engagement behaviours. Specifically, we present our findings for RQ1: How do children engage with a performance art robot, and what causes disengagement in this? Additionally, a thematic analysis of the participants' interview responses aims to build an understanding of children's attitudes towards robots and NED. Through this presentation, we attempt to answer RQ2: What are the children's perceptions of their specific interaction with NED?

## 6.1 Interaction Videos

The participants' behaviours were analysed over 4 hours of interaction videos. Three themes emerged from this analysis: 1) Initiating and maintaining engagement, 2) Lack of robot expressivity and reciprocity, and 3) Unmet expectations.

The first theme explores the different ways participants approached and tried to communicate with NED, such as the range of utilised interaction modalities and strategies to maintain NED's engagement. The second theme presents NED's ability to respond to participants, and limitations of the choregraphed motions. The third theme reviews the impact of expectation on the quality of the interaction, and how failing to meet that expectation can lead to disengagement. These themes form the basis of the discussion from an engagement and disengagement perspective. For the remainder of this paper, individual participants and their quotations will be referred to and labelled as **P1** – **P18**.

Initiating and Maintaining Engagement. The most common modalities for initiating engagement with NED were gesture type actions, such as waving and spreading out the arms. Out of 18 participants, 17 were observed performing some form of gesture to initiate engagement with NED. The second most common modality was voice, with participants often greeting NED or asking NED to do something. Example quotes from the participants include: "Hello!", "Good morning!" – **P3**, "NED do the dance!" – **P5**. Four participants were observed attempting to talk to NED. Interestingly, these modalities for initiating engagement were observed when NED was in the states: Sleep, and Search. Participants easily recognized when NED was in active engagement with them, as all participants either looked at or referred to the end effector during the Interact state. Upon successful engagement, participants employed one or a combination of interaction strategies. The following interaction strategies were observed in the data: Follow (13), Copy (11), Dance (10), and Play (8), where the number in parentheses denotes the total number of participants for whom this strategy was observed.

Lack of Robot Expressivity or Reciprocity. Twelve out of the eighteen participants expressed some form of discontent with NED's inability to do anything beyond following them, despite this being the intended programmed motion for the robot. During the coding process, the following codes emerged: Confused (10), Discouraged (7), Disappointed (6), Disinterested (1), Fed up (3), and Frustrated (5), where the number in parentheses denotes the total number of participants the code was observed for.

The robot state Sleep was observed to negatively affect most participants. Throughout the video annotations, appearance of the Sleep state would often be followed by codes related to discontent, disappointment, or frustration. Examples of participants' frustrations include: "It should just let me do all the work!" – **P3**, "How does this guy get tired all the time?" – **P3**, "Well he's [NED] not starting, so I can't play with him." – **P9**, "Why does it keep going to sleep?" – **P10**.

Unmet Expectations. The naming of the robot, the Never-Ending Dancer, likely influenced the participants' preconceptions about the robot's capabilities, several participants referred to "dance" being in the robot's name. This name implies that 1) NED dances, and 2) it dances forever, as suggested by the phrase "Never-Ending".

Addressing the first implication, we find that the definition of dance is subjective [24]. However, in total, 10 participants attempted some form of dance with NED, and 9 out of 10 exhibited visible confusion, disappointment, or discouragement when NED did not adequately respond or reciprocate specific dance movements. For instance, **P8** performed a dance routine, which involved quickly rolling one closed fist over another, reminiscent of a 1970's dance style. However, being a single robotic arm, NED was unable to reciprocate this style of dancing. Moreover, NED failed to show any responsiveness or difference in motion to suggest that it had understood **P8's** actions. As a result, **P8** stood in front of NED motionless with a discouraged look on their face.

The second implication of the name "Never-Ending Dancer" is when the robot performs no motion when not interacting with a person, or when the system fails to detect a person, as was the case with the first 5 participants at the National Festival of Making in Blackburn. The definition for never-ending is absolute, meaning "to have or seeming to have no end" [25]. Eleven out of 18 participants showed an almost immediate negative response to the robot entering the Sleep state. This effect was prominent during the first 2 minutes of interactions. The strong negative reaction to the Sleep state suggests that when users are actively engaging with the system, and the system fails to effectively convey or reciprocate their engagement, users may perceive this behaviour as negative or ineffective. Resulting in disengagement from the interaction.

### 6.2 Interview Data

When asked "What do you think NED is?", 12 participants identified NED as a robot of some kind, "A Robot." – **P1** and **P3**, "A dancing robot." – **P8**, "A robot with long arms… and a dress!" – **P13**. When asked to explain why they thought NED was a robot, the majority of comments referred to the upper portion of NED, i.e., the Niryo NED robotic arm. In response to the question "How do you know it's a robot?", **P2** replied "Because he's got mechanical things on him.", and when asked to clarify what they meant, **P2** stated "Like his head.". This suggests that participants were confidently able to identify which parts of NED were robotic. However, some participants identified NED as "A bird dancing." – **P4**, "He looks kind of like a bird." – **P10**, referencing the costume, feathers, and end effector gripper appearing like a bird's beak, "that hand [gripper] looks like a beak." – **P9**. Throughout the study NED was referred to as "it", to reduce potential bias. However, most of the participants identified NED as a "he", likely due to the traditionally masculine association of the name Ned, derived from the name Edward or similar variations.

When asked about NED's purpose, "What do you think NED does?", participants answered positively that it either dances, follows, copies, or plays with you. "Dance?" – **P1**, "I think he will move around depending on where you are." – **P9**, "He's a copy-cat." – **P8**, "He dances and plays with you." – **P4**. However, a few participants had more cynical ideas about what NED does. "It's got this camera inside… He takes…spies on you." – **P5**. Although, when asked how they felt about NED, **P5** responded, "A bit excited.", suggesting that despite **P5** alluding to NED having ulterior motives, the experience was positive overall due to the uncertainty of NED's behaviour.

Eleven participants were under the impression that NED was capable of thought or feeling. Participants reported a variety of emotions that NED might be feeling, "…if he was a person, he would like be a bit confused? …Because he's like standing on a post and he's not allowed to go off…" – **P10**, "I think he likes me, but then when I kind of look at him, he looks away from me, so I don't really know." – **P16**. "Happy." – **P2**. Conversely, some participants understood that NED was "programmed" and therefore could not think or feel, "Well AI, even if he was AI, he wouldn't be able to think of anything." – **P9**. There was no statistical difference between the ages of those that agreed NED was capable of thought or feeling, and those who did not. However, participants that did not think NED could think were more likely in the interview to reference popular content creators who specialise in "Robotics", "DIY creations", and science channels, suggesting a preconception of robotics and translating that understanding to the reality of NED. Three participants did not respond to the question.

The interviews also included questions about what participants liked and what additional features they would like NED to have in the future. Twelve participants enjoyed the way NED moved, by following, copying, dancing, or playing. Three participants made positive comments about NED's costume, and 1 participant commented on the type of music that accompanied NED. The most requested feature for future implementations was voice, or voice recognition, with 4 participants expressing a desire to talk to NED or ask it questions, such as "What's his favourite football team?" – **P5**. In general, participants suggested improving NED's ability to move, indicating that if it could copy or follow better, or move around the room, they would have enjoyed interacting with NED more. This suggests multimodal interaction strategies, may be a valid approach for the enhanced engagement of audience members.

Children were able to competently identify NED as a robot, primarily due to its mechanical appearance, particularly its "head". They believed NED's purpose was to engage in playful activities such as dancing and copying. While some children attributed human-like thoughts and feelings to NED, others, influenced by their knowledge of robotics, recognised its programmed nature. The children's overall experience with NED was positive, this was reflected in their enjoyment of its movements and their suggestions for future enhancements, particularly in terms of interactivity and mobility. All 18 participants confirmed that they had fun with NED and 17 participants indicated that they would return to Thingamabobas in the future. Only 1 participant was unable to finish the interview due to time constraints, while 1 other participant could not complete the interview due to losing interest.

## 7 Conclusion

In this paper, we report the results of an in-the-wild experiment of 18 children's interactions with an autonomous robotic system in a performative art setting. We investigate how these children initiate and maintain engagement and the possible causes for disengagement. Our findings suggest that the children that participated in the installation were capable of recognising and engaging with a robotic art performer and are motivated through curiosity to maintain that engagement.

Our observations highlight the importance of designing HRI systems in the performative arts domain with careful consideration of the audience's capabilities, perceptions, and expectations. While we identify that children exhibit enthusiasm and curiosity towards a robotic art performer, critical factors such as the robot's expressiveness, adaptability to audience cues and interaction modalities, influence their engagement.

Our research underscores the significance of addressing these considerations to foster successful interactions between children and robotic art performers. By understanding and accommodating the audiences' perceptions and expectations, future designs can mitigate the risk of failed interactions and enhance overall engagement. This study contributes valuable insights into optimising HRI systems within the performative arts context, for creating engaging and meaningful experiences, especially for young audiences.

## 8 Limitations

Several limitations were identified in this study. Firstly, the sample size and diversity could be improved for better generalisability, as the study involved only 18 participants from specific events, potentially skewing the sample towards families interested in technology and art. The study did not extensively explore the socio-demographic backgrounds of participants, which could influence children's interactions with robots. Future research should aim for a more diverse participant pool. Additionally, the average interaction time with NED was 4.4 minutes, this may not capture the full range of possible interactions and engagement strategies, suggesting that longer interactions could provide more comprehensive data and reveal new insights.

NED's design, while creatively reinterpreted as a kinetic sculpture, also had several limitations. Its expressivity and reciprocity were limited, leading to frustration and disengagement. The robotic arm's motion capabilities did not meet participants' expectations, and the Sleep state caused negative reactions when the robot appeared unresponsive. These design issues highlight the need for improvements in how states and transitions are programmed. Furthermore, the study primarily used qualitative data analysis, which, while comprehensive, lacks the robustness of quantitative methods. We plan integrate both qualitative and quantitative approaches in future research.

The novelty effect of this system was not measurable, due to the relatively short duration of the interactions. However, the performative nature of the system and the art installation that it is a part of, aims to exploit novelty to capture the interests of audiences and place them into a state of disbelief. Contradictory to the current literature, [14] argues for the reframing of the novelty effect to be a valuable source of information, such that novelty should be an "original feature of an experience". We argue that, in the context of performance arts or similar theatrical domain, where interest, engagement, and disbelief hold great importance, the novelty effect should be a phenomena that is taken into account, and that HRI researchers design for.

## 9 Future Work

The future direction of this work is framed within an existing taxonomy for analysing human-robot failures [26], with suggestions on how to better ensure engagement in CRI. The four distinct failure types are: Design, System, Expectation, and User.

**Design.** The design of the system's motions, or lack of motions, has ultimately resulted in users disengaging from the interaction due to the system appearing not to recognise, understand, or appreciate the user's efforts to engage. Although, some participants displayed signs of resilience and perseverance when confronted with robot state instability, the Sleep state's lack of motion and frequency of initiation negatively affected users and their perceptions of the system. Future iterations should consider removing the Sleep state or inviting audience members to initiate engagement using additional modalities, such as voice, addressing both the first and second themes.

**System.** The reliability of the facial detection algorithm deployed in the original system was inadequate in low-light conditions. As a result, participants were not recognised properly despite often being in an optimal space in front of the robot. This issue compounded the failure of the Sleep state's lack of motion, leading some participants to believe that the system was not or chose not to interact with them. Switching to a more reliable method of audience member detection reduced the frequency of false negatives and provided a more stable interaction, allowing participants to initiate and maintain engagement more easily.

**Expectation.** The user may be falsely led to believe that the system is capable of more than it can manage. As expressed by most of the participants, any future iteration of the system should be improved to increase its expressivity and range or style of motion. Perhaps methods for analysing audience members' motions and replaying via the robot arm could allow users to play with new and interesting ways of engaging with the system. The naming of the system should also be considered carefully, such that the system capabilities do not fall below the user's expectations.

**User.** No instances of user failure were observed, although one unintentional failure did occur when participants left the interaction space. This resulted in the system being unable to detect them and forced it to enter the Sleep state. This had a negative impact on the emotional state and engagement of some participants. Nevertheless, participants quickly understood the interaction boundaries and adapted appropriately. This behaviour may prove beneficial, as it allows users to explore the system independently, ensuring users maintain their agency. The recognition of these behaviours and the provision of autonomous guidance or cues to re-engage users in the interaction could enhance user experiences.

**Acknowledgments.** The authors would like to thank Dr Feng Zhou and Mithun Poozhiyil for their contribution to the implementation at the early stages of development. This work was supported by Lakeside Arts, Nottingham, UK, UKRI Engineering and Physical Sciences Research Council Centre for Doctoral Training in Horizon [grant number EP/S023305/1], UKRI Trustworthy Autonomous Systems Hub [grant number EP/V00784X/1] (TAS RRI), Horizon: Digital Economy Hub at the University of Nottingham [EP/G065802/1] (HoRRIzon), AI UK: Creating an International Ecosystem for Responsible AI Research and Innovation [EP/Y009800/1] (RAKE), and using public funding by Arts Council England.

**Disclosure of Interests.** The authors have no competing interests to declare that are relevant to the content of this paper.